\documentclass[10pt,twocolumn,letterpaper, dvipsnames, usenames, svgnames, x11names ]{article}

\usepackage[pagenumbers]{cvpr} 

\usepackage{graphicx}
\usepackage{amsmath}
\usepackage{amssymb}
\usepackage{booktabs}
\usepackage{longtable}
\usepackage{multirow}
\usepackage{array}
\usepackage{comment}
\usepackage{colortbl}
\usepackage{enumerate}
\usepackage{enumitem}
\usepackage{algorithm}
\usepackage[noend]{algpseudocode}
\usepackage{fontawesome}
\usepackage[dvipsnames]{xcolor}
\usepackage{bbm}
\usepackage{soul}
\setstcolor{red}

\usepackage{xfrac}

\makeatletter
\newcommand*\bigcdot{\mathpalette\bigcdot@{.5}}
\newcommand*\bigcdot@[2]{\mathbin{\vcenter{\hbox{\scalebox{#2}{$\m@th#1\bullet$}}}}}
\makeatother

\newcommand{\name}{\emph{SUNY}}

\usepackage[pagebackref,breaklinks,colorlinks]{hyperref}

\usepackage[capitalize]{cleveref}
\crefname{section}{Sec.}{Secs.}
\Crefname{section}{Section}{Sections}
\Crefname{table}{Table}{Tables}
\crefname{table}{Tab.}{Tabs.}

\def\cvprPaperID{3430} 
\def\confName{CVPR}
\def\confYear{2023}

\begin{document}

\title{\name{}: A Visual Interpretation Framework for \\Convolutional Neural Networks from a Necessary and Sufficient Perspective}

\author{Xiwei Xuan$^{1,*}$ \quad Ziquan Deng$^{1}$ \quad Hsuan-Tien Lin$^{2}$ \quad Zhaodan Kong$^{1}$ \quad Kwan-Liu Ma$^1$  \\
{$^1$University of California, Davis} \quad 
{$^2$National Taiwan University} \quad 
{$^*$Corresponding Author} \quad \\
{\tt\small {\{xwxuan, ziqdeng\}}@ucdavis.edu} \quad
{\tt\small {htlin}@csie.ntu.edu.tw}\quad
{\tt\small {\{zdkong, klma\}}@ucdavis.edu}
}

\maketitle


\begin{abstract}
Researchers have proposed various methods for visually interpreting the Convolutional Neural Network (CNN) via saliency maps, which include Class-Activation-Map (CAM) based approaches as a leading family. However, in terms of the internal design logic, existing CAM-based approaches often overlook the causal perspective that answers the core ``why'' question to help humans understand the explanation. Additionally, current CNN explanations lack the consideration of both necessity and sufficiency, two complementary sides of a desirable explanation. This paper presents a causality-driven framework, SUNY, designed to rationalize the explanations toward better human understanding. Using the CNN model's input features or internal filters as hypothetical causes, SUNY generates explanations by bi-directional quantifications on both the necessary and sufficient perspectives. Extensive evaluations justify that SUNY not only produces more informative and convincing explanations from the angles of necessity and sufficiency, but also achieves performances competitive to other approaches across different CNN architectures over large-scale datasets, including ILSVRC2012 and CUB-200-2011. 

\end{abstract}

\section{Introduction}
\label{sec:intro}
In computer vision, an important research direction is to give Convolutional Neural Network (CNN) more transparency to extend its deployment on a broader basis.
This paper addresses the eXplainable Artificial Intelligence (XAI)~\cite{gunning2017explainable} problem corresponding to CNN for natural image classification, i.e., reasoning why a classifier makes particular decisions.
Specifically, we study a leading family of techniques called Class-Activation-Maps (CAMs), which present heatmaps highlighting image portions associated with a model's class prediction.
\begin{figure}[thbp]
  \centering
   \includegraphics[width=1\linewidth]{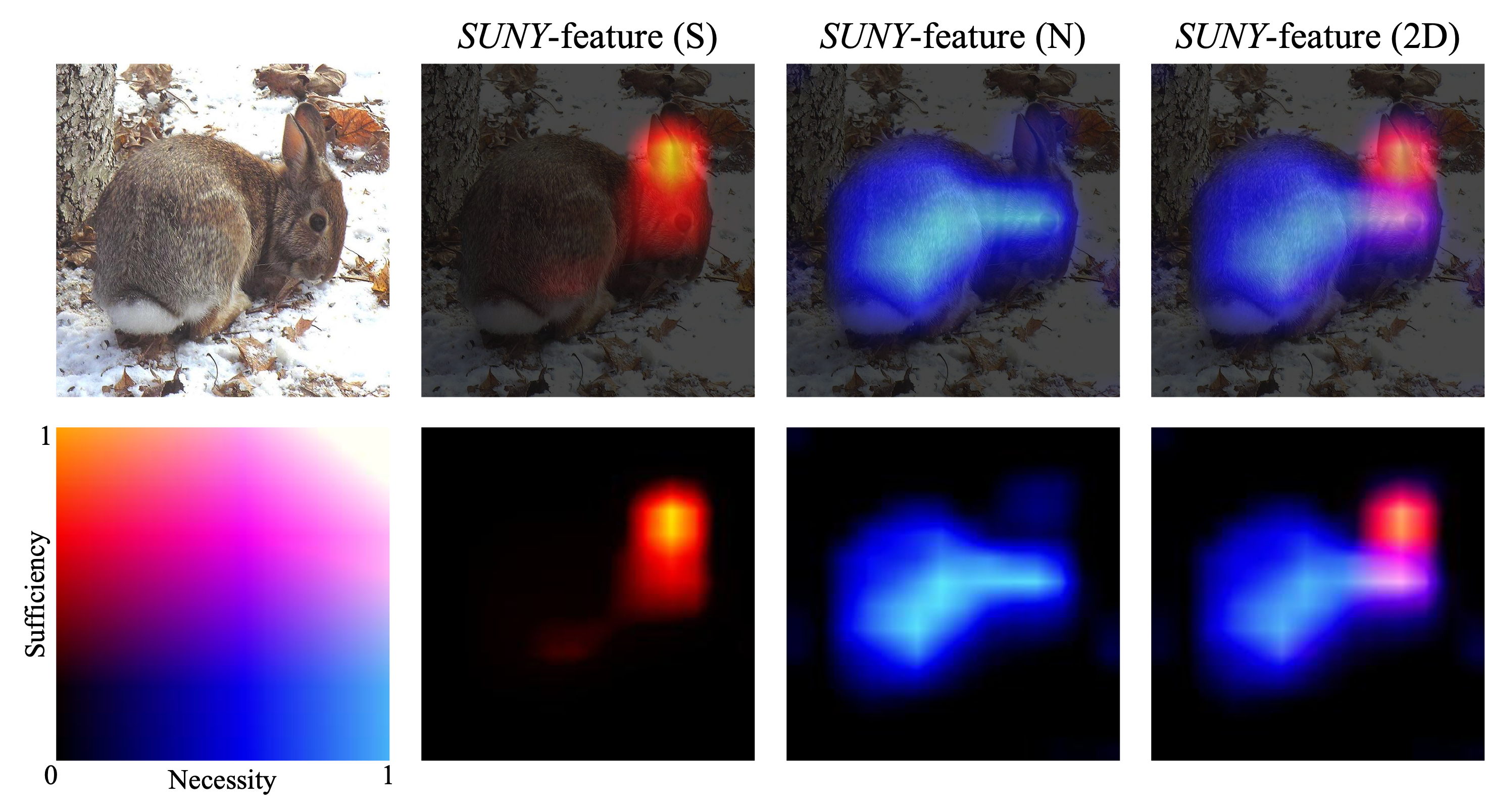}
   \caption{An example of \name{} explanations. \name{} highlights sufficient and necessary input regions w.r.t.\;the model's prediction towards the target class. The 2D saliency map is the first-of-its-kind visual explanation design to the best of our knowledge.}
   \label{fig:Intro Compare}
\end{figure}
The pioneering method in the family is CAM~\cite{zhou2016learning}, which produces saliency maps by linearly combining the convolutional feature maps with weights generated by a global average pooling layer.
CAM's structural restriction on the global average pooling layer prohibits its use for general CNNs, and a series of gradient-weighted CAMs~\cite{selvaraju2017grad,chattopadhay2018grad,omeiza2019smooth} have been proposed to overcome the restriction.
However, such CAMs suffer from gradients' saturation and vanishing issues, which cause the resulting explanations to be noisy~\cite{ghorbani2019interpretation}.
To bypass the shortcomings of gradients, Score-CAM~\cite{wang2019score} and Group-CAM~\cite{zhang2021group} weight the feature maps by contribution scores, referring to the corresponding input features' importance to the model output.

The design of Score-CAM and Group-CAM to measure the model's class prediction (outcome) when only keeping specific input features (cause) matches the idea of causal \textit{sufficiency (S)}, while the other aspect, \textit{necessity (N)}, is missing.
\textit{N} refers to changing the hypothetical causes and measuring the outcome differences, and \textit{S} investigates the outcome stability when specific causes are unchanged.
Both are desirable and essential perspectives for a successful explanation~\cite{lipton1990contrastive,woodward2006sensitive,grinfeld2020causal,watson2021local}.
Other CAMs, on the other hand, lack both causal semantics in their design, which makes it challenging to answer the core ``why'' questions and match human instincts.

Producing a saliency map through causal explanations requires calculating the importance of each cause (e.g., a region of input image or a model filter) when interacting with other causes.
Techniques for quantifying the importance of causes consist of two main genres, \textit{causal effect}~\cite{glymour2016causal} and \textit{responsibility}~\cite{chockler2004responsibility}.
\textit{Causal effect} measures the outcome differences when intervening on the causes; \textit{responsibility} measures the minimal set of causes needed to change the outcome.
In \textit{causal effect} analysis, the effect of a group of causes is a single value, which cannot be easily attributed to distinguish the importance of each individual cause.
\textit{Responsibility} also fails to provide a decent importance score for a single cause because it measures the number of causes rather than the actual outcome differences.
That is, these two causal quantification genres cannot be directly taken to produce a sound saliency map.
Furthermore, both \textit{causal effect} and \textit{responsibility} typically focus on $N$ and overlook $S$.
There are a few works addressing CNN causal explanations by adopting either of these two methods, such as CexCNN~\cite{debbi2021causal} and~\cite{harradon2018causal}.
However, the aforementioned limitations have not yet been carefully addressed.
To summarize, current CNN explanations, to some extent, fail in (1) human interpretability (not designed from a causal perspective, like GradCAM), (2) precision (failing to generate decent importance for individual cause, given their basis of \textit{causal effect} or \textit{responsibility}), or (3) completeness (missing either \textit{N} or \textit{S} perspective, like CexCNN).
Therefore, it calls for more research efforts in this demanding field of CNN explanations to fulfill the deficiencies.

Considering the issues mentioned above, we propose an explanation framework called \textbf{SU}fficiency and \textbf{N}ecessit\textbf{Y} causal explanation (\name{}), which interprets the CNN classifier by analyzing how the cooperation of a single cause and other causes affect the model's decision.
\name{} regards either model filters or input features as the hypothesized causes and quantifies each cause's importance towards the class prediction from angles of both \textit{N} and \textit{S}.
Specifically, we draw on the strength of both \textit{causal effect} and \textit{responsibility} to propose \textit{N-S responsibility}, which solves the three issues above by (1) being designed from causality theory that is in line with instinctive human habits; (2) providing importance for each individual cause when interacting with other causes by quantizing the effect on top of responsibility; (3) conducting a bi-directional quantification from both \textit{N} and \textit{S} perspectives.
Based on the qualitative evaluation, including the semantic evaluation and the sanity check, we demonstrate that \name{} provides a more faithful and interpretable visual explanation for CNN models.
Comprehensive experiment evaluations on benchmark datasets, including ILSVRC2012~\cite{ILSVRC15} and CUB-200-2011~\cite{welinder2010caltech}, validate that \name{} outperforms other popular saliency map-based visual explanation methods. 
Our key contributions are summarized as follows: 
\begin{itemize}
\setlength{\itemsep}{1pt}
\setlength{\parsep}{1pt}
\setlength{\parskip}{1pt}
\item  We define \textbf{N-S Shapley Values} for a bi-directional causal importance quantification for CNNs from \textit{N} and \textit{S} perspectives.
\item  We introduce a flexible framework, \name{}, which can provide valid CNN visual explanations by analyzing either model filters or input features.
\item We provide 2D saliency maps for a more informative visual explanation, which is the first 2D visual explanation design of CNN to the best of our knowledge.
\item Extensive quantitative and qualitative experiments highlight our explanations' effectiveness and uncover the potential of \name{} in real-world problem-solving.
\end{itemize}

\section{Related Work}
\label{sec:related}
\noindent\textbf{Saliency Map Explanations for CNNs.}
The class-discri-minative visual explanation is a key tool for CNN interpretation by highlighting important regions corresponding to a model's particular prediction.
Two widely-adopted types are perturbation-based methods~\cite{zeiler2014visualizing,ribeiro2016should,petsiuk2018rise} and CAMs~\cite{zhou2016learning,selvaraju2017grad,chattopadhay2018grad,smilkov2017smoothgrad,wang2019score,zhang2021group}.
Perturbation-based methods such as RISE~\cite{petsiuk2018rise} are inspired by a human-understandable concept of measuring the output when the inputs are perturbed, which is in line with the \textit{N} causal semantics.
Somehow enumerating and examining all possible input perturbations is computationally challenging~\cite{ivanovs2021perturbation}, making such methods infeasible for applications demanding fast explanations.

CAMs compute a weighted sum of feature maps to generate explanations, which are often more computationally efficient.
CAM~\cite{zhou2016learning} is the pioneering method in this type, which requires a global average pooling layer in the CNN architecture.
GradCAM~\cite{selvaraju2017grad} and a series of its variations, such as Grad-CAM++~\cite{chattopadhay2018grad} and SmoothGrad~\cite{smilkov2017smoothgrad}, address this structural restriction by using gradients as weights, hence applicable to a broader variety of CNN families.
However, gradients knowingly suffer from vanishing issues and introduce noise to the saliency map~\cite{ghorbani2019interpretation}.
To fix the gradient issues, Score-CAM~\cite{wang2019score} and Group-CAM~\cite{zhang2021group} weight feature maps by their corresponding input features' contribution to the output, which matches the \textit{S} perspective of causal semantics and is implicitly consistent with the ideas of perturbation-based methods. 

\noindent\textbf{Causal Importance Evaluations.}
\label{subsec:current_res}
Based on the HP causality~\cite{pearl2009causality, halpern2020causes} introduced by Halpern and Pearl, recent research for causal importance quantification involves two main genres, \textit{causal effect} and \textit{responsibility}.
The average \textit{causal effect}~\cite{glymour2016causal} measures the expected outcome difference when the values of selected causes are changed, which regards the set of causes as a whole.
An individual cause can thus affect the outcome very differently, i.e., receive completely different effect quantification, when being analyzed with different sets of interacting causes~\cite{yao2021survey}.
The second genre, the degree of \textit{responsibility}~\cite{chockler2004responsibility}, quantifies the minimal set of causes whose removal leads to a satisfactory outcome change, which overlooks the actual value of the outcome~\cite{lagnado2013causal,baier2021verification}.
Besides, current \textit{causal effect} and \textit{responsibility} both focus on the \textit{N} perspective, i.e., the change of causes, and the other \textit{S} perspective needs to be fulfilled for a thorough explanation~\cite{lipton1990contrastive,woodward2006sensitive,grinfeld2020causal,watson2021local}.
More recent works such as \cite{watson2021local} and \cite{kommiya2021towards} address this deficiency by measuring the causal importance from both \textit{N} and \textit{S}.
However, their proposed quantification focuses on the probability of the outcome change when causes are changed (\textit{N}) or unchanged (\textit{S}), which also misses the actual outcome value.

\noindent\textbf{Causal Explanations for CNNs.}
Given recent works on CNN causal explanations, two main groups exist according to the type of the hypothesized causes -- the model's input features or inner filters.
As for the first group, \cite{chattopadhyay2019neural} and \cite{harradon2018causal} estimate the average \textit{causal effect} of input features.
The Shapley value-inspired methods~\cite{vstrumbelj2014explaining,datta2016algorithmic,lundberg2017unified,sundararajan2020many,aas2021explaining}  also implicitly analyze the causality of input features from the \textit{S} perspective. However, these methods are too computationally intensive or only provide a hazy outline of salient regions.
Two CAM-based methods, Score-CAM~\cite{wang2019score} and Group-CAM~\cite{zhang2021group}, as discussed before, can be regarded as partially causal-grounded by analyzing the \textit{sufficiency} of input features.
On the other side, the perturbation-based methods ~\cite{zeiler2014visualizing,ribeiro2016should,petsiuk2018rise} analyze input features' \textit{necessity}.

The second group of methods interprets CNN through the model filter intervention.
For example, \cite{narendra2018explaining} explores CNN's inner workings by ranking filters from a layer according to their counterfactual (CF) influence.
However, without any visualizations, this causal explanation of CNN is less intuitive than saliency maps.
CexCNN~\cite{debbi2021causal} provides visualizations with the help of CAMs and uses \textit{responsibility} to quantify the CNN's filter importance, which means only the \textit{N} aspect has been analyzed.

\section{CNN Classification as a Causal Mechanism}
\label{sec:motivation}
With a CNN natural image classifier $M$, an input image $I$, and a target class $c$, our goal is to provide saliency maps to highlight which image portions lead to the model's prediction of $c$ from a causal perspective.
First, we recall the definition of \textit{causal mechanism} as follows.

\noindent\textbf{Causal mechanism~\cite{falleti2009context}.}
A portable concept explaining how and why the hypothesized causes, in a given context $u$, contribute to a particular outcome. 

Next, we define the outcome and causes in our problem, then demonstrate how we establish the causal mechanism and formulate our problem based on it.

\noindent\textbf{Outcome: }
As a class-discriminative explanation, we directly measure the model's prediction probability w.r.t.\;the target class and set it as the outcome.

\noindent\textbf{Causes: }
Given that existing CNN causal explanations regard input features or model filters as causes, we allow the flexibility of setting either of them as causes.
Specifically, a single cause is (1) a region of the image identified by a feature map; or (2) a model filter in a convolutional layer.

When considering input features as causes, the context $u$ is an unknown process determining the combination of image features.
Regarding model filters as causes, the context $u$ is the image region on which we apply the model filters.
After establishing the causal mechanism, we can quantify the effect of causes by counterfactually intervening on the hypothesized causes and measuring the outcome, whose results can be used as weights for feature maps' linear combination.
Finally, we can produce saliency maps as a causal explanation for the CNN classifier.

\section{The Proposed Approach}
\label{sec:solution}
\begin{figure*}[htbp]
  \centering
   \includegraphics[width=1.0\linewidth]{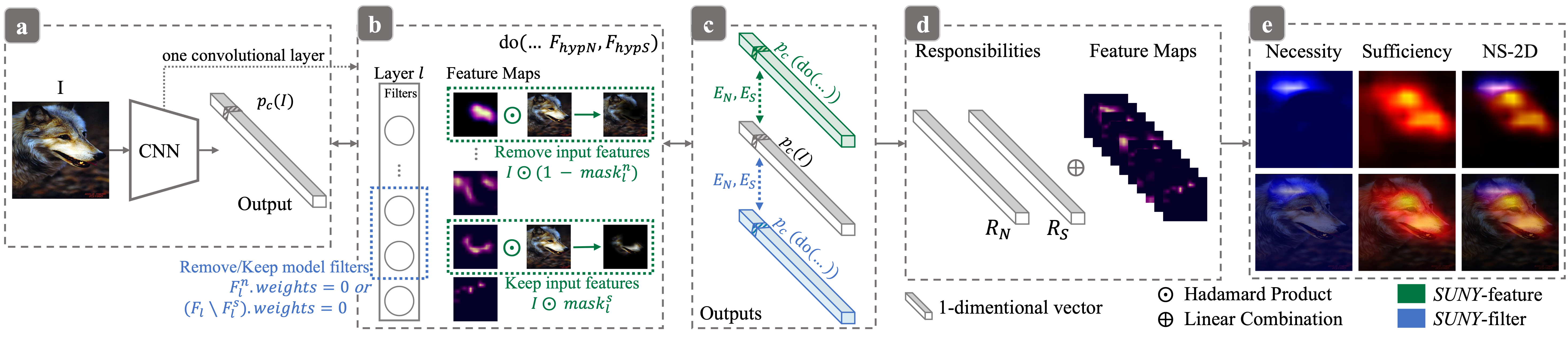}
\caption{
Overview of \name{} framework.
Phase(a) is a forward pass of input image $I$ through a CNN model, where the prediction probability of the target class is $p_{c}(F)$.
Phase(b)-(e) present the generation of \textcolor{ForestGreen}{\name{}-feature (green)} and \textcolor{RoyalBlue}{\name{}-filter (blue)}, respectively, referring to different types of hypothesized causes.
Note that they are not simultaneous processes.
In Phase(b), we obtain filters and feature maps of a specified layer, and intervene on model filters or the corresponding input features.
We get new prediction probabilities after the intervention and calculate \textbf{N-S Effect}, $E_N$, $E_S$ in Phase(c), which are fed back to Phase(b) to construct hypothesized cause sets $F_{hypN}$ and $F_{hypS}$.
Through intervening on $F_{hypN}$ and $F_{hypS}$ (Phase(b)), we can obtain $E_N$, $E_S$ (Phase(c)) and \textbf{N-S Responsibilities} $R_N$ and $R_S$ (Phase(d)), which are weights for the linear combination of feature maps.
The saliency maps are generated in Phase(e), where we show \name{}-feature results as an example.
Implementation details are included in Sec.~\ref{sec:solution}.
}
   \label{fig:framework}
\end{figure*}

\subsection{Causal Importance Evaluation}
\label{subsec:ns_res}

To address the deficiencies of current causal importance evaluations as discussed in Sec.~\ref{sec:related}, our evaluation should follow the below design requirements:

\begin{enumerate}[start=1,label={\bfseries R\arabic*},leftmargin=*,topsep=2px,partopsep=2px]
\setlength{\itemsep}{0pt}%
\setlength{\parskip}{0pt}
    \item \label{req:set_size}
Our method should measure the importance of each individual cause in a group of coordinating causes. 
    \item \label{req:dual_direction}
Our method should quantify the actual effect towards the outcome and pay attention to both \textit{N} and \textit{S}.
\end{enumerate}

The Shapley value's framework for assessing marginal contributions across various coalitions lays the groundwork for achieving~\ref{req:set_size}. 
Additionally, we further define the necessity and sufficiency value functions to accomplish~\ref{req:dual_direction}.
However, previous SHAP~\cite{lundberg2017unified} image analyses segment input images into equally-sized patches, restricting finer distinctions. 
Our model-integrated method regards a single feature map (or a set of feature maps) as a cause $f_i$ (or a set of causes $F_*$), thereby providing more granular explanations. 
We utilize a general formulation of Shapley values in the following definition.

For a set of causes (i.e., a set of feature maps) to be analyzed as $F_*$, we define \textit{\textbf{N}} value function to measure the degree of the outcome change when removing $F_*$:
\begin{equation}\label{eqn:n_eff}
   E_N(F_*) = \frac{p_{c}(I)-p_{c}({\rm do} \; (F \setminus F_*))} {p_{c}(I)},
\end{equation}
where ${\rm do}(F \setminus F_*)$ represents the CF intervention of removing $F_*$.
And $p_{c}(\bigcdot)$ refers to the model's prediction probability w.r.t.\;a target class $c$, where $p_{c}(I)$ is its original value without any intervention; $p_{c}({\rm do}(F \setminus F_*))$ represents the value after the removal intervention.
A higher $E_{N}(F_*)$ indicates a more \textit{necessary} $F_*$.
Similarly, \textit{\textbf{S}} value function is defined as:
\begin{equation}\label{eqn:s_eff}
   E_S(F_*) = \frac{p_{c}({\rm do}\; (F_*))} {p_{c}(I)},
\end{equation}
where ${\rm do}(F_*)$ represents the CF intervention of only keeping $F_*$, i.e., removing $\{F \setminus F_*\}$.
A higher $E_{S}(F_*)$ indicates a more \textit{sufficient} $F_*$.

To fulfill~\ref{req:set_size}, different from covering all elements, we tend to focus on more necessary (sufficient) ones. 
$\forall f_i \in F$, where $f_i$ is a single cause, we set $F_* = \{f_i\}$ to calculate $E_N(f_i)$ ($E_S(f_i)$) and construct a set $F_N \subseteq F$ ($F_S \subseteq F$) by combining the relatively more necessary (\textit{sufficiency}) $f_i$.
Then to analyze a single cause $f_n \in F_N$, we calculate the $N$ Shapley value as:

\begin{equation}\label{eqn:n_res}
\begin{split}
   R_N(f_n) =  \sum_{F^\prime\subseteq \{F_N \setminus f_n\}}\frac{|F^\prime|!(|F_N|-|F^\prime|-1)!}{|F_N|!}\\
   \times[E_N(F^\prime \cup {f_n})- E_N(F^\prime)].
\end{split}
\end{equation}
Similarly, we can calculate $S$ Shapley value for $f_s\in F_S$ as:
\begin{equation}\label{eqn:s_res}
\begin{split}
    R_S(f_s)=\sum_{F^\prime\subseteq \{F_S \setminus f_s\}} \frac{|F^\prime|!(|F_S|-|F^\prime|-1)!}{|F_S|!}\\
    \times[E_S(F^\prime \cup {f_s})- E_S(F^\prime)].
\end{split}
\end{equation}
In the implementation, we reduce the amount of computation by estimating Eqns.(\ref{eqn:n_res}), (\ref{eqn:s_res}) using Shapley sampling values method~\cite{vstrumbelj2014explaining}.
Additionally, for $f_n^\prime \in \{F\setminus F_N\}$ and $f_s^\prime \in \{F\setminus F_S\}$, we set $R_N(f_n^\prime) = 0$ and $R_S(f_s^\prime) = 0$.

\begin{figure*}[htbp]
  \centering
   \includegraphics[width=1\linewidth]{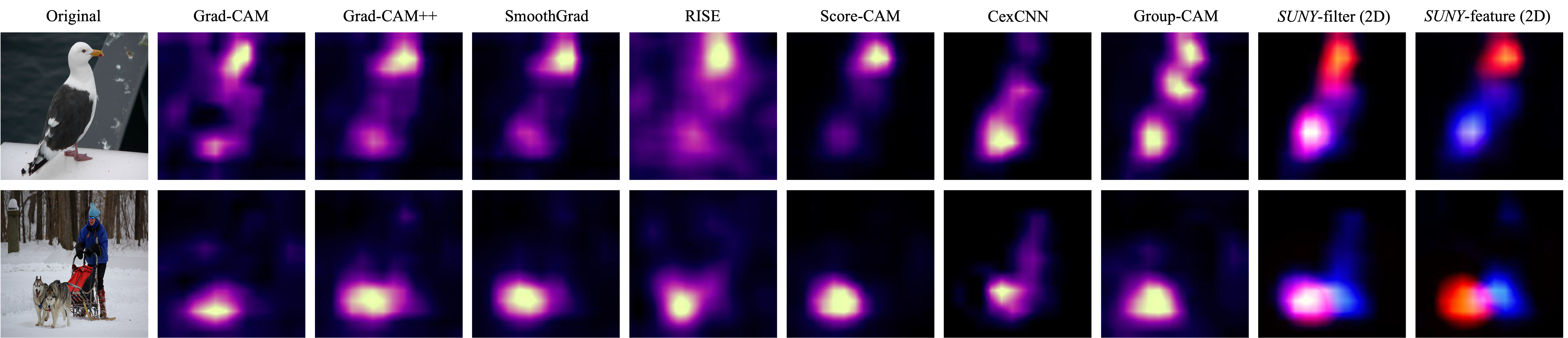}
\caption{Visual comparison of saliency maps from different methods. The first row: a VGG16 trained on CUB-200-2011, and the image is correctly predicted as {\fontfamily{qcr}\selectfont Gull}. The second row: a VGG16 trained on ILSVRC2012, and the image is correctly predicted as {\fontfamily{qcr}\selectfont Dog-sled}.
}
   \label{fig:Compare all}
\end{figure*}

\subsection{\textbf{\name{}} Implementation}

We present the implementation of \name{} in Alg.~\ref{alg:suny} and a visual overview in Fig.~\ref{fig:framework}.
Following the definitions in Sec.~\ref{subsec:ns_res}, \name{} provides causality-driven CNN visual explanations regarding input features or model filters as hypothesized causes, respectively, represented by the cause type $E$ in Alg.~\ref{alg:suny} with values \textit{``feature''} or \textit{``filter''}.
The corresponding explanations are \name{}-feature and \name{}-filter.
\begin{algorithm}[bhpt]
\caption{\name{}: Causal Explanation of CNN}\label{alg:suny}

\begin{algorithmic}[1]
  \Require Image $I$, model $M$, layer $l$, class $c$, cause type $E$
  \Ensure N-S saliency maps: $N_{map}$, $S_{map}$
\State{$p_{c}(\bigcdot) =$ Softmax$(M(\bigcdot))[c]$}\Comment{prediction probability on $c$}
\State{$A_l \gets M_l(I)$} \Comment{ feature maps of the layer $l$}
\State{$F_{N}, F_{S} \gets $ getHypCauses($I, M, l, E$)}
\State{$R_N, R_S \gets zeros(A_l.shape[0])$}\Comment{initialize responsibilities}
\If {$E$ is $``feature"$}
    \For{$A_l^n$ \textbf{in} $F_{N}$}
    \State{$mask \gets$ norm(upsample$(A_l^n)$)}
    \State{Compute $R_N(A_l^n)$ based on Eqn.(\ref{eqn:n_res})}
    \EndFor
    \For{$A_l^s$ \textbf{in} $F_{S}$}
    \State{$mask \gets$ norm(upsample$(A_l^s)$)}
    \State{Compute $R_S(A_l^s)$ based on Eqn.(\ref{eqn:s_res})}
    \EndFor
\ElsIf {$E$ is $``filter"$}
    \For{$F_l^n$ \textbf{in} $F_{N}$} 
    \State{$M^{n} \gets$ pruneFilters$(M, F_l^n)$}
    \State{$p^{n}_{c}(\bigcdot) =$ Softmax$(M^{n}(\bigcdot))[c]$}
    \State{Compute $R_N(F_l^n)$ based on Eqn.(\ref{eqn:n_res})}
    \EndFor
    \For{$F_l^s$ \textbf{in} $F_{S}$} 
    \State{$M^{s} \gets$ pruneFilters$(M, (F_l \setminus F_l^s))$}
    \State{$p^{s}_{c}(\bigcdot) =$ Softmax$(M^{s}(\bigcdot))[c]$}
    \State{Compute $R_S(F_l^s)$ based on Eqn.(\ref{eqn:s_res})}
    \EndFor
\EndIf
\State{$N_{map} =$ norm(upsample(Relu$(\sum {R_N}^{i}A_l^i)$))}
\State{$S_{map} =$ norm(upsample(Relu$(\sum {R_S}^{i}A_l^i)$))}
\State{\Return{$N_{map}$, $S_{map}$}}
\end{algorithmic}
\end{algorithm}
We first define $p_c(\bigcdot)$ as a function to calculate the model's prediction probability w.r.t.\;a class $c$ for the input denoted by $\bigcdot$, as shown in line 1 of Alg.~\ref{alg:suny}.
Next, in line 3, we construct $F_{N}$ and $F_{S}$, where $F_{N}$ is the set of hypothetical single causes $f_n$ with relatively higher $E_N(f_n)$.
Similarly, $F_{S}$ contains all hypothetical single causes $f_s$ with higher $E_S(f_s)$.
We then calculate \textbf{N-S Shapley Values} for every single cause in $F_{N}$ and $F_{S}$ following lines 5 - 20.
Specifically, for \name{}-feature (lines 6 - 11), we upsample and normalize the feature maps, $A_l^n$ and $A_l^s$, and use the generated $mask$ as a feature extractor to intervene on input features from the image $I$.
The intervention ${\rm do}(F \setminus F_*)$ (removing $F_*$) is realized by the Hadamard product, $(I \bigodot (1 - mask))$.
Similarly, ${\rm do}(F_*)$ (keeping $F_*$) is implemented as $(I \bigodot mask)$.
\name{}-filter (lines 13 - 20) removes and keeps the hypothesized causes by filter pruning, which means setting the corresponding filters' weights to zero.
Line 14 and line 18 correspond to ${\rm do}(F \setminus F_l^n)$ and ${\rm do}(F_l^s)$, respectively.
After calculating \textbf{N-S Shapley Values} for all single causes, in lines 21 and 22, we obtain small saliency maps by $Relu(\sum {R_N}^{i}A_l^i)$, $Relu(\sum {R_S}^{i}A_l^i)$ and then upsample and normalize them to get the final saliency maps.
The operation $norm$ represents the min-max normalization w.r.t.\;each single map, $norm(X) = \frac{X - min(X)}{max(X) - min(X)}$, and $upsample$ represents the bilinear interpolation.

In the following section, we validate the performance of \name{} from both quantitative and qualitative perspectives.

\section{Experiments}
\label{sec:experiments}
\subsection{Experimental Setup}
\begin{figure*}[htbp]
  \centering
   \includegraphics[width=1\linewidth]{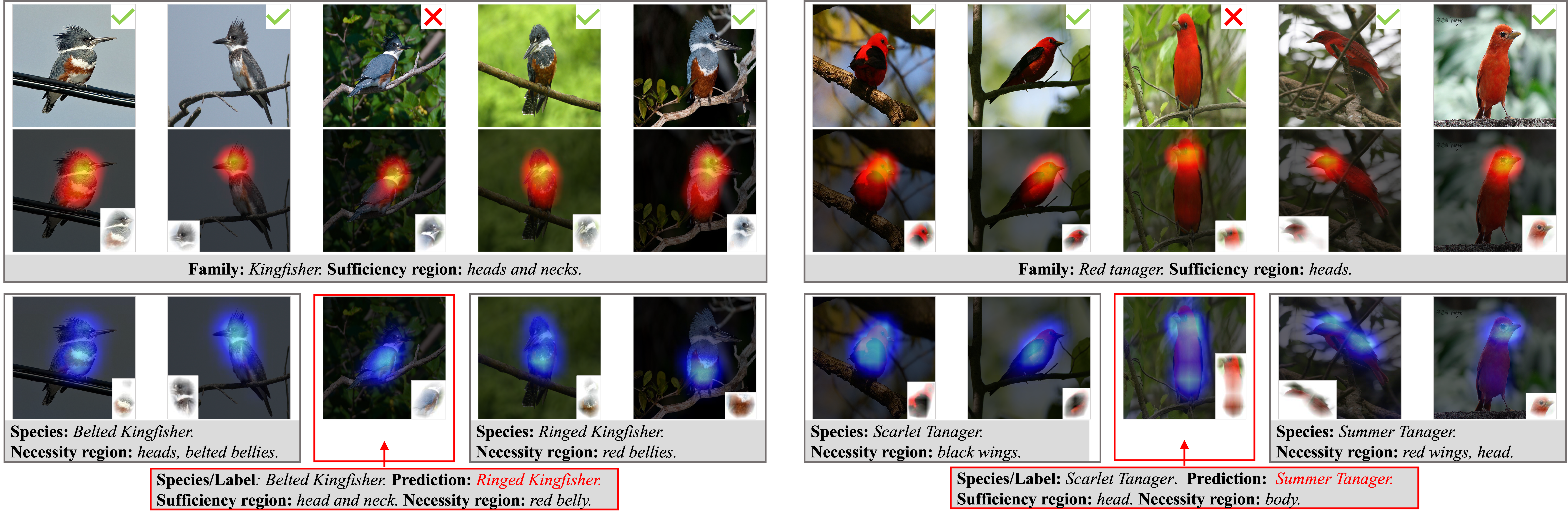}
\caption{
Semantic evaluation of \name{} explanations for a VGG16 trained on CUB for bird species classification.
The bird images in the first row are from four bird species belonging to two families and the correct/incorrect predictions are marked by~\textcolor{LimeGreen}{\faCheck} and~\textcolor{red}{\faTimes}, respectively.
For the two images marked by~\textcolor{red}{\faTimes}, the model mistakes the actual species with the other species under the same family.
Each column corresponds to one image; the second and third rows: \textit{sufficiency} and \textit{necessity} heatmaps.
The small image in the bottom corner of each heatmap presents the highlighted image portion.}
\label{fig:Bird Causal}
\end{figure*}

\noindent \textbf{Baseline Methods.}
The baseline methods we select to compare should be localization approaches.
Similar to \name{}, they follow two requirements: (1) class-discriminative (different explanations for different specified classes); (2) presenting heatmaps highlighting input regions. 
Specifically, we compare \name{} with seven popular methods proposed between 2017 and 2021, covering approaches that require white-box models or black-box models: Grad-CAM~\cite{selvaraju2017grad}, Grad-CAM++~\cite{chattopadhay2018grad}, SmoothGrad~\cite{smilkov2017smoothgrad}, RISE~\cite{petsiuk2018rise}, Score-CAM~\cite{wang2019score}, CexCNN~\cite{debbi2021causal}, and Group-CAM~\cite{zhang2021group}.

\noindent \textbf{Datasets and CNN Models.}
The experiments involve the validation sets of two commonly-used image datasets, ILSVRC2012 (ILSVRC)~\cite{ILSVRC15} with $1000$ classes and $50k$ images and CUB-200-2011 (CUB)~\cite{welinder2010caltech} with $200$ classes and $5794$ images.
We use all methods to explain three CNNs with different architectures, including VGG16~\cite{simonyan2014very}, Inception-v3~\cite{szegedy2016rethinking}, and ResNet50~\cite{he2016deep}.

\noindent \textbf{Implementation Details.}
We implement \name{} and the seven aforementioned visual explanation methods in Python using PyTorch framework~\cite{paszke2017automatic}.
Specifically, we run the official or publicly available code of other methods as the results at the same data scale are unavailable.
The platform is equipped with two NVIDIA RTX 3090 GPUs. 
Unless explicitly stated, the comparisons discussed in this section are conducted following the same settings, including --

\noindent(1) Images are converted to the RGB format, resized to $224\times224$ (VGG16) or $299\times299$ (Inception-v3), transformed to tensors, and normalized to the range of $[0,1]$.

\noindent(2) All visual explanation methods' results are generated for both models in the validation sets of both datasets, where we use all images predicted correctly by the models.

\noindent (3) Explanations are applied to the last convolutional layer for explaining the predicted class. We bilinearly interpolate the results to the required size of each experiment.

\subsection{Semantic Evaluation}

In Fig.~\ref{fig:Compare all}, we visually compare \name{} with other saliency map explanations and observe two advantages: 
(1) Saliency maps provided by \name{} contain fewer noises.
(2) \name{} uniquely provides both \textit{necessary} and \textit{sufficient} information to support interpretation. For example, for the first image, \name{} tells that the bottom wing is \textit{necessary} and the head is \textit{sufficient} for {\fontfamily{qcr}\selectfont Gull} prediction. For the second image, \name{} shows that the sled is \textit{necessary} and the dog is \textit{sufficient} for the prediction of {\fontfamily{qcr}\selectfont Dog-sled}.
However, other explanations overlook one side of \textit{necessity} or \textit{sufficiency}, and none provide such distinction.
More examples can be found in the Supplementary Materials.

\noindent \textbf{Causal explanations with \name{}.}
As shown in Fig.~\ref{fig:Bird Causal}, we provide images from CUB corresponding to four bird species in two families\footnote{Family is a higher-level taxonomic category than species~\cite{enwiki:1116635228}}.
By inspecting the \textit{sufficiency} heatmaps in the second row, we can observe that the highlighted regions are similar for those belonging to the same bird family. -- All {\fontfamily{qcr}\selectfont kingfisher} images share similar heads and belted necks; all {\fontfamily{qcr}\selectfont red tanager} images share similar heads.
This explains why the model correctly identifies the bird family for every image.
From the third row, we find the \textit{necessity} heatmaps provide further explanations through the following observations.
For the {\fontfamily{qcr}\selectfont kingfisher} family, the belted bellies are highlighted in images predicted as {\fontfamily{qcr}\selectfont belted kingfisher}, while the red bellies are highlighted in the images predicted as {\fontfamily{qcr}\selectfont ringed kingfisher}.
This explains why the third image is mistaken -- the red belly is easily observable through this view and is identical to a {\fontfamily{qcr}\selectfont ringed kingfisher}.
For the {\fontfamily{qcr}\selectfont red tanager} family, the black wings are highlighted in images predicted as {\fontfamily{qcr}\selectfont Scarlet tanager}, while the red wings or heads are highlighted in the images predicted as {\fontfamily{qcr}\selectfont summer tanager}. For the {\fontfamily{qcr}\selectfont Scarlet tanager} image being misclassified, the black wing feature is not observable from this view, and the head and red body are captured by the model, which are characteristics belonging to the other species.
Therefore, Fig.~\ref{fig:Bird Causal} shows that \textit{sufficiency} and \textit{necessity} provides semantically-complementary explanations to better support model behavior interpretation.
More examples are provided in the Supplementary Materials.

\begin{table*}[ht]
\centering
\resizebox{\textwidth}{!}{
\begin{tabular}{p{1.2cm}<{\centering}| p{3cm}<{\centering}| p{1.5cm}<{\centering} p{1.6cm}<{\centering} p{1.5cm}<{\centering}| p{1.5cm}<{\centering} p{1.6cm}<{\centering} p{1.5cm}<{\centering} |p{1.5cm}<{\centering} p{1.6cm}<{\centering} p{1.5cm}<{\centering}}
\toprule[0.8pt]
                          &                           & \multicolumn{3}{c|}{VGG 16}    & \multicolumn{3}{c|}{Inception\_v3} & \multicolumn{3}{c}{ResNet50}\\
\multirow{-2}{*}{Dataset} & \multirow{-2}{*}{Methods} & Deletion $\downarrow$ & Insertion $\uparrow$ & Overall $\uparrow$ & Deletion $\downarrow$ & Insertion $\uparrow$  & Overall $\uparrow$ & Deletion $\downarrow$ & Insertion $\uparrow$  & Overall $\uparrow$ \\ \hline
                          &Grad-CAM\cite{selvaraju2017grad}                 &  0.1098          & 0.6112          &  0.5015        & 0.1276          &0.6567           & 0.5291       & 0.1796 & 0.6889 &0.5093 \\
                          &Grad-CAM++\cite{chattopadhay2018grad}               & 0.1155         & 0.6033           & 0.4878        & 0.1309         & 0.6476          & 0.5167     &0.1847 &0.6799 & 0.4952      \\
                          & SmoothGrad \cite{smilkov2017smoothgrad}              & 0.1136         &0.6023           &0.4887        & 0.1317         &  0.6465         & 0.5148   &0.1849 &0.6800 & 0.4951       \\
                          & RISE \cite{petsiuk2018rise}                       &0.1185           & 0.6188        & 0.5003         & 0.1404          &0.6444     &  0.5040          &\textcolor{Green}{0.1303} &0.6932 &\textcolor{blue}{0.5629}\\
                          & Score-CAM \cite{wang2019score}              
                          & 0.1070         &  0.6382         & 0.5312        
                          & 0.1309         & 0.6528          & 0.5219           & 0.2319 & 0.6218 & 0.3898\\
                          & CexCNN \cite{debbi2021causal}              
                          & 0.1161         & 0.6025          &0.4864         
                          & 0.1355         & 0.6543          & 0.5188           & 0.1886 &0.6443 &0.4557\\
                          & Group-CAM \cite{zhang2021group}              
                          &0.1138          & 0.6218          & 0.5080        
                          &0.1292          & 0.6545          & 0.5253          & 0.1794 & 0.6904 & 0.5110\\
 &
  \cellcolor[HTML]{EFEFEF}\name{}-filter (Ours) & 
  \cellcolor[HTML]{EFEFEF}{\textcolor{blue}{0.1019}} &
  \cellcolor[HTML]{EFEFEF}{\textcolor{blue}{0.6389}} &
  \cellcolor[HTML]{EFEFEF}{\textcolor{blue}{0.5370}} &
  \cellcolor[HTML]{EFEFEF}{\textcolor{blue}{0.1258}}  &
  \cellcolor[HTML]{EFEFEF}{\textcolor{blue}{0.6570}}  &
  \cellcolor[HTML]{EFEFEF}{\textcolor{blue}{0.5312}} &
  \cellcolor[HTML]{EFEFEF}{0.1425} &
  \cellcolor[HTML]{EFEFEF}{\textcolor{blue}{0.6957}} &
  \cellcolor[HTML]{EFEFEF}{0.5532}
  \\
\multirow{-9}{*}{ILSVRC} &
  \cellcolor[HTML]{EFEFEF}\name{}-feature (Ours) &
  \cellcolor[HTML]{EFEFEF}{\textcolor{Green}{0.1005}} &
  \cellcolor[HTML]{EFEFEF}{\textcolor{Green}{0.6468}} &
  \cellcolor[HTML]{EFEFEF}{\textcolor{Green}{0.5462}} &
  \cellcolor[HTML]{EFEFEF}{\textcolor{Green}{0.1215}} &
  \cellcolor[HTML]{EFEFEF}{\textcolor{Green}{0.6603}} &
  \cellcolor[HTML]{EFEFEF}{\textcolor{Green}{0.5388}} &
  \cellcolor[HTML]{EFEFEF}{\textcolor{blue}{0.1323}} &
  \cellcolor[HTML]{EFEFEF}{\textcolor{Green}{0.6988}} &
  \cellcolor[HTML]{EFEFEF}{\textcolor{Green}{0.5665}} \\ \hline
                          & Grad-CAM\cite{selvaraju2017grad}                 &  0.0558        &  \textcolor{Green}{0.7617}         &  \textcolor{blue}{0.7059}       & 0.0963         & 0.7323          & 0.6360          & 0.0930 &0.6452 &0.5522 \\
                          & Grad-CAM++\cite{chattopadhay2018grad}               &  0.0589        & 0.7541           & 0.6951        & 0.0950         &  0.7281          & 0.6331     & 0.0972 &0.6407 & 0.5434     \\
                          & SmoothGrad \cite{smilkov2017smoothgrad}              & 0.0594         & 0.7489          & 0.6895         & 0.0977         & 0.7244          & 0.6266     & 0.0974 & 0.6405 &0.5431     \\
                          & RISE \cite{petsiuk2018rise}              &0.0560          & 0.7583          &  0.7023       & 0.0855         & 0.7168          & 0.6314        & \textcolor{blue}{0.0570}  &0.6567 & \textcolor{blue}{0.5996} \\
                          & Score-CAM\cite{wang2019score}               &  \textcolor{blue}{0.0542}        & 0.7575          &   0.7033      &0.0901          & 0.7326          &  0.6424          &0.0995 &0.6351 &0.5355\\
                        & CexCNN \cite{debbi2021causal}              & 0.0630         & 0.7389          & 0.6760        & 0.1017         & 0.7283          &  0.6267         & 0.1014 & 0.6173 & 0.5159 \\
                          & Group-CAM \cite{zhang2021group}              & 0.0606         & 0.7521          & 0.6915        & 0.0971         & 0.7290          & 0.6318       & 0.0926 & 0.6458 & 0.5532  \\
 &
  \cellcolor[HTML]{EFEFEF}\name{}-filter (Ours) &
  \cellcolor[HTML]{EFEFEF} 0.0544&
  \cellcolor[HTML]{EFEFEF} 0.7556&
  \cellcolor[HTML]{EFEFEF} 0.7012&
  \cellcolor[HTML]{EFEFEF} {\textcolor{blue}{0.0853}}&
  \cellcolor[HTML]{EFEFEF} {\textcolor{blue}{0.7333}}&
  \cellcolor[HTML]{EFEFEF} {\textcolor{blue}{0.6480}}
  & \cellcolor[HTML]{EFEFEF} 0.0682
  & \cellcolor[HTML]{EFEFEF} {\textcolor{Green}{0.6613}} 
  & \cellcolor[HTML]{EFEFEF} 0.5931\\
\multirow{-9}{*}{CUB} &
  \cellcolor[HTML]{EFEFEF}\name{}-feature (Ours) &
  \cellcolor[HTML]{EFEFEF}{\textcolor{Green}{0.0518}} &
  \cellcolor[HTML]{EFEFEF}{\textcolor{blue}{0.7591}} &
  \cellcolor[HTML]{EFEFEF}{\textcolor{Green}{0.7073}} &
  \cellcolor[HTML]{EFEFEF}{\textcolor{Green}{0.0842}}&
  \cellcolor[HTML]{EFEFEF}{\textcolor{Green}{0.7361}} &
  \cellcolor[HTML]{EFEFEF}{\textcolor{Green}{0.6519}} 
  & \cellcolor[HTML]{EFEFEF}{\textcolor{Green}{0.0562}}
  & \cellcolor[HTML]{EFEFEF}{\textcolor{blue}{0.6645}}
  & \cellcolor[HTML]{EFEFEF}{\textcolor{Green}{0.6083}}\\  \bottomrule[0.8pt]
\end{tabular}}
\caption{Comparative evaluation w.r.t.\;the \textit{deletion}, \textit{insertion}, and \textit{overall} AUC on ILSVRC and CUB with VGG16, Inception\_v3, and ResNet50, where lower \textit{deletion}, higher \textit{insertion}, and higher \textit{overall} indicate a better explanation. 
The \textcolor{Green} {first} and \textcolor{blue} {second} best performances are marked in \textcolor{Green} {green} and \textcolor{blue} {blue}, respectively.}
\label{tab:AUC}
\end{table*}

\subsection{Necessity and Sufficiency Evaluation}
\label{subsec:NS_eval}

\noindent \textbf{Deletion and Insertion.}
To evaluate the saliency regions' effects on the model's prediction, we conduct the \textit{deletion and insertion} experiments proposed in~\cite{petsiuk2018rise}.
According to the saliency map, the \textit{deletion} metric quantify the model's predicted probability upon the removal of more and more important image pixels, while the complementary \textit{insertion} metric measure the same variable as the image pixels adds back from the most important to unimportant ones.
Consistent with~\cite{petsiuk2018rise, zhang2021group}, we utilize the Area Under the probability Curve (AUC) to quantify the results, with lower \textit{deletion}, higher \textit{insertion}, and higher \textit{overall(insertion-deletion)} indicative of a better explanation.
We follow the same experimental settings as~\cite{zhang2021group}, including the blurring method and the step size to remove/add pixels. 
The comparison results are presented in Table~\ref{tab:AUC},
where we can observe that except for VGG16 on CUB, both \name{}-feature and \name{}-filter outperform other approaches in terms of \textit{deletion}, \textit{insertion} and \textit{overall} AUC scores. 
In the case of VGG16 on CUB, \name{}-feature still outperforms most of the other explanations, and \name{}-filter achieves competitive results.

\noindent \textbf{N-S Quantification.}
We introduce the \textit{N-S Quantification} metric to quantify the average \textit{N} and \textit{S} degree of a visual explanation method.
The \textit{N-S scores} correspond to one input image $I$ are given by $N_{score}=\frac{p_{c}(I)-p_{c}(I\bigodot(1 - map))}{p_{c}(I)\times Size_{map}}$, $S_{score}=\frac{p_{c}(I\bigodot(map))}{p_{c}(I)\times Size_{map}}$, respectively,
where $map$ is the saliency map provided by the explanation method, $p_{c}(\bigcdot)$ is the model's prediction probability w.r.t.\;the target class $c$, and $Size_{map}=\frac{\sum_{i, j}(1_{true}map[i, j] \neq 0)}{map.h\times map.w}$.
\begin{figure}[bhtp]
  \centering
   \includegraphics[width=1\linewidth]{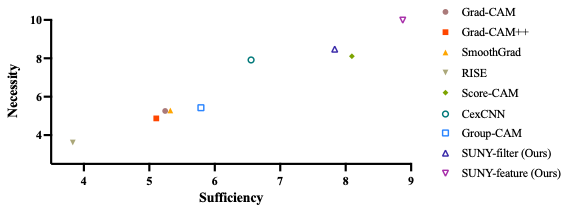}
   \caption{Comparison of \name{} with seven visual explanation methods in terms of the \textit{N-S Quantification} metric.}
   \label{fig:NSCUB}
\end{figure}
As the highlighted image regions being removed or kept, $N_{score}$ and $S_{score}$ measure how \textit{necessary} or \textit{sufficient} these regions are for the model's decision, with larger scores indicative of higher \textit{N/S}.
For every visual explanation method, we conduct the \textit{N-S Quantification} with VGG16 on the correct-predicted images in the CUB validation set, and calculate the average $N_{score}$ and $S_{score}$ as its \textit{necessary} and \textit{sufficient} level.
The result is visualized in Fig.~\ref{fig:NSCUB}, which highlights that \name{} outperforms most of the seven methods in the ability to identify the most \textit{necessary} and \textit{sufficient} image regions corresponding to the model's class prediction.

\subsection{Saliency Attack}
Researchers have proposed a series of local adversarial attack approaches~\cite{dong2020robust, wang2022attention, xiang2021local, dai2022saliency} guided by saliency maps such as CAM~\cite{dong2020robust,wang2022attention}, Grad-CAM~\cite{xiang2021local}, and others~\cite{dai2022saliency}, which is to fool CNN models by perturbing a small image region highlighted by saliency maps.
These methods require the saliency maps to be ``minimal and essential''~\cite{dai2022saliency}, i.e., be capable of capturing image portions that most significantly impact the model output.
Inspired by current saliency attack approaches, we propose an evaluation metric, $Attack_{score}=\frac{FlipRate}{AvgAttackSize}$, to validate whether \name{} explanations can detect the most important regions w.r.t.\;the model's decision.

Specifically, we add random Gaussian noise to the saliency regions $I'=(I + noise \bigodot map)$, and then check whether this operation can change the model's decision, i.e., $Flip = 1$ if ${argmax(p(I))}\neq{argmax(p(I'))}$, where $p(\bigcdot)$ is the model's prediction probability.
The $FlipRate$ represents the ``essential'' level of the saliency regions.
To validate whether the region is ``minimal'', we include $AvgAttackSize$, which is the average size of all saliency maps. 
For a single saliency map, the calculation of $Size_{map}$ is the same as shown in Sec.~\ref{subsec:NS_eval}.
Finally, we can obtain the $Attack_{score}$ and the comparison is reported in Table~\ref{tab:Attack}, which proves that bi-directional explanations provided by \name{} are better at highlighting the most important image region corresponding to the model's decision.

\begin{table}[ht]
\centering
\resizebox{\columnwidth}{!}{
\begin{tabular}{p{1.2cm}<{\centering}|p{3cm}<{\centering}|p{1.45cm}<{\centering} p{1.45cm}<{\centering}p{1.45cm}<{\centering}}
\toprule[0.8pt]
                          &                                            & \multicolumn{3}{c}{Saliency $Attack_{score}$ $\uparrow$}                          \\
\multirow{-2}{*}{Dataset} & \multirow{-2}{*}{Methods}                  & VGG16                    & Inception\_v3      & ResNet50       \\ \hline
                           & Grad-CAM\cite{selvaraju2017grad}                  & 0.9615        &   1.0435         & 0.7674     \\
                          & Grad-CAM++\cite{chattopadhay2018grad}                & 0.9991        & 0.9821      & 0.8751        \\
                          & SmoothGrad\cite{smilkov2017smoothgrad}               &  1.0449       & 0.9675    & 0.8776           \\
                          & RISE\cite{petsiuk2018rise}               
                          & 0.9928        &  0.7353       & 1.0259       \\
                          & Score-CAM\cite{wang2019score}               & 0.5326        & 0.9673      & 0.3378     \\
                          & CexCNN\cite{debbi2021causal}                       & 1.6341         & 1.0653    &0.6393 \\
                          & Group-CAM\cite{zhang2021group}                      & 1.1556          &  1.0200   & 0.8020\\
                          & \cellcolor[HTML]{EFEFEF}\name{}-filter (Ours) & \cellcolor[HTML]{EFEFEF}{\textcolor{blue}{1.6602}} & \cellcolor[HTML]{EFEFEF}{\textcolor{blue}{1.5725}}  & \cellcolor[HTML]{EFEFEF}{\textcolor{blue}{1.3228}}\\
\multirow{-9}{*}{ILSVRC} & \cellcolor[HTML]{EFEFEF}\name{}-feature (Ours) & \cellcolor[HTML]{EFEFEF}{\textcolor{Green}{2.0452}} & \cellcolor[HTML]{EFEFEF}{\textcolor{Green}{1.9874}} 
& \cellcolor[HTML]{EFEFEF}{\textcolor{Green}{1.5619}}\\ \hline
                          & GradCam\cite{selvaraju2017grad}                 & 0.5969        & 0.5985   & 0.4694             \\
                          & GradCam++\cite{chattopadhay2018grad}                & 0.6670        &  0.5950   & 0.5257           \\
                          & SmoothGrad\cite{smilkov2017smoothgrad}                & 0.7783        & 0.5929    & 0.5260           \\
                          & RISE \cite{petsiuk2018rise}              
                          & 0.5063        & 0.3860    & 1.1286           \\
                          & Score-CAM\cite{wang2019score}               
                          & 1.2215        &  0.5989    & 0.8027          \\
                          & CexCNN\cite{debbi2021causal}               
                          &  1.2673       & 0.5898      & 0.4171         \\
                          & Group-CAM\cite{zhang2021group}                
                          & 0.6742        & 0.5951       & 0.4991        \\
                          & \cellcolor[HTML]{EFEFEF}\name{}-filter (Ours) & \cellcolor[HTML]{EFEFEF}{\textcolor{blue}{1.3691}} & \cellcolor[HTML]{EFEFEF}{\textcolor{blue}{0.8867}} & \cellcolor[HTML]{EFEFEF}{\textcolor{blue}{1.2885}}\\
\multirow{-9}{*}{CUB}    & \cellcolor[HTML]{EFEFEF}\name{}-feature (Ours) & \cellcolor[HTML]{EFEFEF}{\textcolor{Green}{2.8111}} & \cellcolor[HTML]{EFEFEF}{\textcolor{Green}{1.0475}}  & \cellcolor[HTML]{EFEFEF}{\textcolor{Green}{1.7747}}\\  \bottomrule[0.8pt]
\end{tabular}
}
\caption{
Comparative evaluation w.r.t.\;the saliency attack score (higher is better) on the validation sets of ILSVRC and CUB with VGG16, Inception\_v3, and ResNet50. The \textcolor{Green} {first} and \textcolor{blue} {second} best performances are marked in \textcolor{Green} {green} and \textcolor{blue} {blue}, respectively.)
}
\label{tab:Attack}
\end{table}

\begin{table}[ht]
\centering
\resizebox{\columnwidth}{!}{
\begin{tabular}{p{1.2cm}<{\centering}|p{3cm}<{\centering}|p{1.45cm}<{\centering} p{1.45cm}<{\centering} p{1.45cm}<{\centering}}
\toprule[0.8pt]
                          &                                            & \multicolumn{3}{c}{Proportion ($\%$) $\uparrow$}                          \\
\multirow{-2}{*}{Dataset} & \multirow{-2}{*}{Methods}                  & VGG16                    & Inception\_v3    & ResNet50         \\ \hline
                           &Grad-CAM\cite{selvaraju2017grad}                  &  57.68  &  66.35      & 59.84        \\
                          & Grad-CAM++\cite{chattopadhay2018grad}               &  61.31       &  65.93    &61.74         \\
                          & SmoothGrad\cite{smilkov2017smoothgrad}               &  62.18       &  65.78  &61.75            \\
                          & RISE\cite{petsiuk2018rise}              
                          & 58.93        &  59.26    &59.48          \\
                          & Score-CAM\cite{wang2019score}               &  64.25      &  65.94    &66.72          \\
                          & CexCNN\cite{debbi2021causal}               
                          &  65.24        & 66.33   &57.39      \\
                          & Group-CAM\cite{zhang2021group}              
                          & 62.70         & 66.17   &60.68      \\
                          & \cellcolor[HTML]{EFEFEF}\name{}-filter (Ours) & \cellcolor[HTML]{EFEFEF}{\textcolor{Green}{65.63}} & \cellcolor[HTML]{EFEFEF}{\textcolor{Green}{67.02}}  & \cellcolor[HTML]{EFEFEF}{\textcolor{Green}{68.64}}\\
\multirow{-9}{*}{ILSVRC} & \cellcolor[HTML]{EFEFEF}\name{}-feature (Ours) & \cellcolor[HTML]{EFEFEF}{\textcolor{blue}{65.61}} & \cellcolor[HTML]{EFEFEF}{\textcolor{blue}{66.71}} & \cellcolor[HTML]{EFEFEF}{\textcolor{blue}{68.02}}\\ \hline
                          & Grad-CAM\cite{selvaraju2017grad}                 & 43.06        & 40.05      &39.02        \\
                          & Grad-CAM++\cite{chattopadhay2018grad}                & 45.45        & 40.45  &41.25             \\
                          & SmoothGrad\cite{smilkov2017smoothgrad}               & 47.12        & 40.34   &41.28            \\
                          & RISE\cite{petsiuk2018rise}                       & 37.28           & 34.74  &36.32 \\
                          & Score-CAM\cite{wang2019score}               
                          &  49.68       &  40.67     &\textcolor{blue}{47.42}        \\
                          & CexCNN\cite{debbi2021causal}               
                          &  37.13        &   41.38   &41.22   \\
                          & Group-CAM\cite{zhang2021group}              
                          & 43.53         &  41.08   &40.36     \\
                          & \cellcolor[HTML]{EFEFEF}\name{}-filter (Ours) & \cellcolor[HTML]{EFEFEF}{\textcolor{Green}{51.74}} & \cellcolor[HTML]{EFEFEF}{\textcolor{Green}{42.03}} & \cellcolor[HTML]{EFEFEF}{\textcolor{Green}{49.74}}\\
\multirow{-9}{*}{CUB}    & \cellcolor[HTML]{EFEFEF}\name{}-feature (Ours) & \cellcolor[HTML]{EFEFEF}{\textcolor{blue}{49.97}} & \cellcolor[HTML]{EFEFEF}{\textcolor{blue}{41.96}}  & \cellcolor[HTML]{EFEFEF}{43.21}\\ \bottomrule[0.8pt]
\end{tabular}
}
\caption{
Comparative evaluation w.r.t.\;the proportion in energy-pointing games (higher is better) on the validation sets of ILSVRC and CUB with VGG16, Inception\_v3, and ResNet50. The \textcolor{Green} {first} and \textcolor{blue} {second} best performances are marked in \textcolor{Green} {green} and \textcolor{blue} {blue}, respectively.
}
\label{tab:Loc}
\end{table}

\subsection{Localization Evaluation}
In this section, we conduct the localization evaluation with the energy-based pointing game~\cite{wang2019score}, a variant of the pointing game~\cite{petsiuk2018rise}.
The goal is to measure the localization ability of saliency maps using the ground-truth bounding box of the target class, $bbox$.
The input image is binarized with $bbox$ by assigning the inside and outside regions with $1$ and $0$, respectively. 
Then, we apply the Hadamard product between the binarized input and the saliency map, the summary of which can quantify how much ``energy'' falls into $bbox$.
The performance is measured by $Proportion=\frac{\sum map[i,j]_{(i,j)\in bbox}}{\sum map[i,j]}$.
For convenience, we consider all correctly-predicted images with single $bbox$ in the validation set of ILSVRC, CUB, respectively, and report the average proportion value for every method in Table~\ref{tab:Loc}. 
The results indicate that both \name{}-filter and \name{}-feature outperform other methods in terms of localization ability by ranking first and second in the energy-based pointing game.

\begin{figure}[htbp]
  \centering
   \includegraphics[width=1\linewidth]{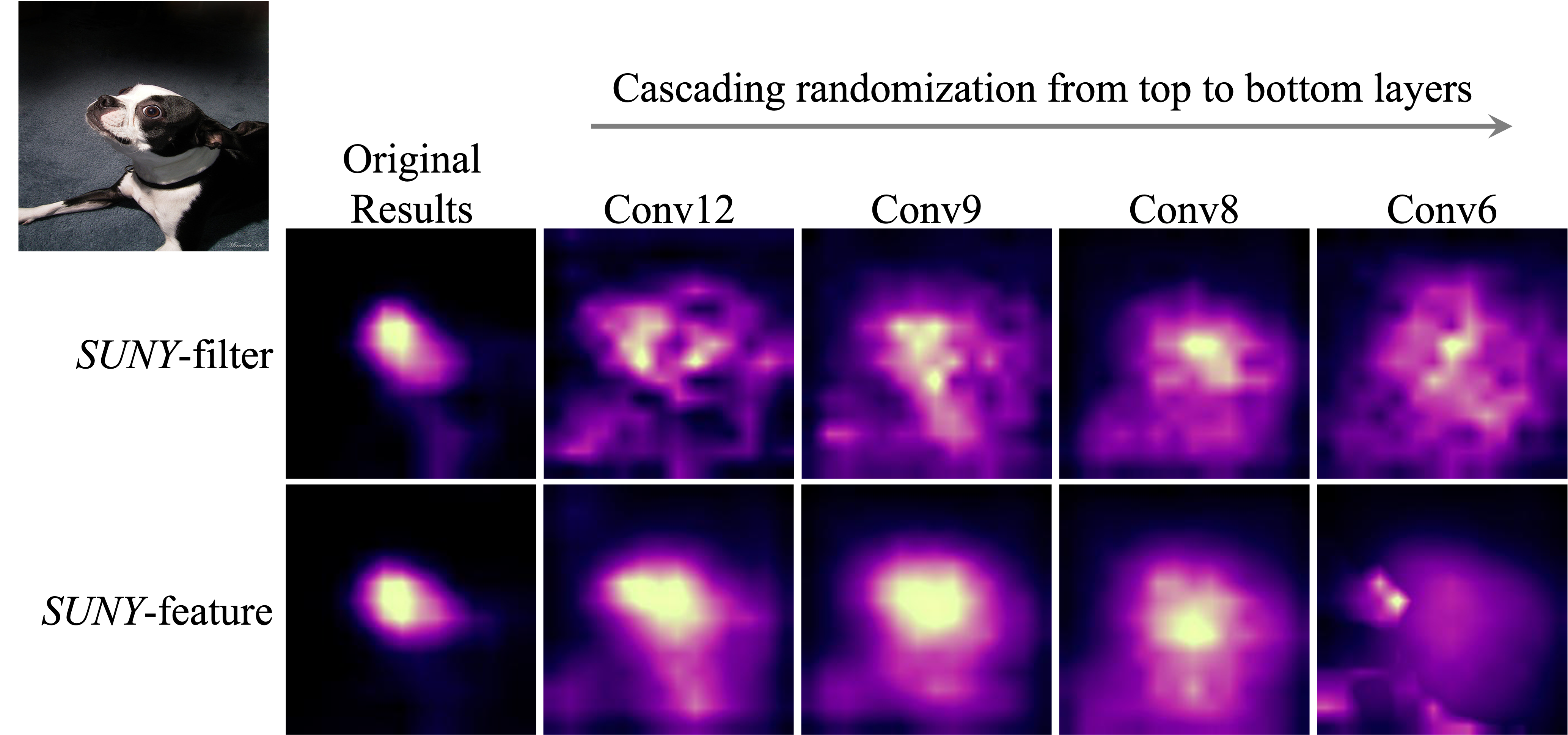}
   \caption{Sanity check of \name{} explanations for VGG16 by cascading randomization with an example image (shown on the top left) from ILSVRC.
   Results of \name{}-filter and \name{}-feature are represented in the first and second row, respectively.
   The first column is the original saliency maps, and the following columns show results after randomizing specific layers.}
   \label{fig:sanity}
\end{figure}

\subsection{Sanity Check}
The sanity check~\cite{adebayo2018sanity} is to validate whether a visual explanation method can be considered as a reliable explanation to reflect the model’s behavior.
We conduct cascading randomization to the model's weights from the top to the bottom layer successively and generate explanations every time after the randomization.
If the saliency maps remain similar for the resulting model with widely differing parameters, it means the method fails the sanity check.
Fig.~\ref{fig:sanity} presents the sanity check of \name{}-filter and \name{}-feature, which indicates significant changes for both explanation results after the model parameter randomization.
Therefore, the explanations provided by \name{} pass the sanity check.

\section{Conclusion}
\label{sec:conclusion}
We design a causality-driven framework, \name{}, for bi-directional CNN interpretation from a \textit{necessary} and \textit{sufficient} perspective.
By establishing a causal mechanism for the CNN classification process, we conduct causal analysis by regarding the CNN model's prediction probability of a target class as the outcome and the model filters or input features as the hypothesized causes, respectively.
Furthermore, our explanations, \name{}-filter and \name{}-feature, are provided by 2D saliency maps, a unique visual explanation design to support a more informative model interpretation.
Our semantic evaluation results show that \name{} can provide more insightful visualizations and demonstrate how model interpretations can benefit from \textit{necessity} and \textit{sufficiency} information.
Moreover, \name{} explanations also pass the sanity check and quantitatively outperform seven other visual explanation methods in necessity and sufficiency evaluation, saliency attack, and localization evaluations using two different large-scale public datasets for two CNN models.

In the future, we plan to exploit the generality of \name{} framework in the scenario of interpreting image segmentation and vision-language tasks. 
In addition, we plan to apply \name{} in data and model validation.

\section*{Acknowledgement}
This work is supported in part by NIBIB under grant no. P41 EB032840. H.-T. Lin is partially supported by the Ministry of Science and Technology in Taiwan via MOST 111-2918-I-002-006 and 112-2628-E-002-030.

{\small
\bibliographystyle{ieee_fullname}
\bibliography{egbib}
}

\end{document}